# Ant Colony Optimization for Inferring Key Gene Interactions

Khalid Raza* and Mahish Kohli
*Department of Computer Science, Jamia Millia Islamia (Central University),*
New Delhi-110025, India
*kraza@jmi.ac.in

**Abstract:** Inferring gene interaction network from gene expression data is an important task in systems biology research. The gene interaction network, especially key interactions, plays an important role in identifying biomarkers for disease that further helps in drug design. Ant colony optimization is an optimization algorithm based on natural evolution and has been used in many optimization problems. In this paper, we applied ant colony optimization algorithm for inferring the key gene interactions from gene expression data. The algorithm has been tested on two different kinds of benchmark datasets and observed that it successfully identify some key gene interactions.

## 1. Introduction

Gene regulatory network (GRN) prediction is an important problem in systems biology. A pre-knowledge of gene interaction network, which are targeting to suppress or express genes, is very handy in drug design for various lethal diseases (such as cancer). Gene interaction network prediction is also important for prediction of salvage pathways. No gene has independent expression, as it is always affected by other genes whose expression interferes with its expression. The expression of a gene is a vector sum of all genes input to it. Thus, a relation can be set up between the expression and the interaction of genes, which is nothing but gene regulation network. Microarray data has become a benchmark source for the inference of gene regulatory networks. Gene network inference from microarray data has been an area of interest, where few people are interested in finding the clusters or the sub networks within a network while some focus on inferring key interactions in a network. Normally, interaction pattern is responsible for expression pattern, say a gene stimulated positively expresses more than the one stimulated negatively. But, in reverse engineering approach, we try to predict the regulation (interaction) pattern from the expression [1][2].

## 2. Review of related works

Researchers worked upon different type of algorithms and approaches for inferring gene interaction network, such as Boolean networks, Bayesian networks, differential equations, machine learning and evolutionary algorithms. Among all these techniques, differential equation based formalism produces acceptable results but problem is that complexity of the algorithm increases (number of equations increases) as we increase the number of genes. Machine learning algorithms, like artificial neural networks (ANN), are also known to be used for predicting interactions but they are so complex that what is happening inside the algorithm is nothing but a black box, which is not good as a biological point of view [3].

On the other hand, nature based algorithms in comparison to other algorithms are simpler in nature and they have been found to be applied in various biological problems from simplest like alignment of sequences to the complex like protein structure prediction [4]. In past, nature based algorithm like Genetic algorithm have been applied directly for optimization of influence matrix [5][6] and have been applied in cooperation with S system [7]. It has been observed that when these applied in the form of hybrids for the inference of networks, they have known to gain much success compared to when applied alone. It has been observed that in hybrids though computationally expensive but they are good for small scale networks, but when it comes to large networks they decline in performance [3]. In past, ant colony optimization (ACO) algorithm has been applied to several bioinformatics problems like drug designing [8][9] and 2D protein folding [10]. It has also been hybridized with genetic algorithm for multiple sequence alignment [11]. In 2009, He & Hui [12] used the ACO algorithm and proposed two algorithms i) Ant-C (An ant based clustering) and ii) Ant-ARM (An ant based association rule mining). With the help of ACO algorithm, they constructed a fully connected network and then removing the edges having pheromone amount lesser than that of an average threshold were removed and thus clusters have been created. All the three above mentioned algorithms being an optimizing algorithms, hence can be used so as to optimize a gene regulatory matrix. A review of soft computing based approach and evolutionary approach can be found in [13] and [14], respectively. In this paper, we applied ACO algorithm for inferring highly correlated key interactions in GRN.

## 3. Materials and Methods

In 2005, Karaboga [15] gave an interesting idea of artificial ants based algorithm known as Ant colony optimization (ACO) algorithm. Ants are blind, but yet known to find the shortest distance between the food source and there native place. Ants use pheromones laid by the other ants as footmarks to follow. And hence an ant reaches the shortest path by using knowledge gained by the other ants and this in form of an algorithm can be used for optimization problems, including gene interaction network optimization. Ant colony optimization is based on evolutionary algorithm, where artificial ants are used, which generate a population of individuals, each individual corresponding to each tour and finally converge to most optimum individual among the population. The most basic problem solved by the ACO algorithm is Travel Salesmen Problem [16][17]. It is a problem where we have $N$ number of cities and one needs to find a route among all possible routes that covers all the cities only once and finally reach the starting city, thus covering minimum distance. Since, ACO is an optimization algorithm; hence it can be used for the optimization of gene interaction matrix. Some of the basic steps of ACO algorithm are described as follows:

**Steps in algorithm**

1. Setting basic parameters such as number of ants, independent trials, number of tour in each trial, etc.
2. Place all ants at different genes randomly.
3. $\tau_{ij}=0$ (Trail of pheromone is zero in initial step)
4. Tabu list initially = {i} (the present gene)
5. Each ant moves to next gene by probability function

$$p_{ij}^k(t) = \begin{cases} \frac{[\tau_{ij}(t)]^\alpha \cdot [\eta_{ij}]^\beta}{\sum_{k \in allowed_k} [\tau_{ik}(t)]^\alpha \cdot [\eta_{ik}]^\beta} & if\ j \in allowed_k \\ 0 & otherwise \end{cases} \quad (1)$$

Where,

$p_{ij}^k(t)$ = The probability of moving from gene 'i' to 'j' at time 't'

k ϵ {1 to m} where m= number of ants

$\eta_{ij}$ = Correlation between the gene 'i' and 'j' or visibility

α , β = parameters controlling importance of visibility ant trail respectively

6. After N iteration all ants complete tour, the best tour is chosen and pheromone trail is updated.
7. If termination condition is met, tour ends, else tabu list is emptied and cycle continues from step 2.

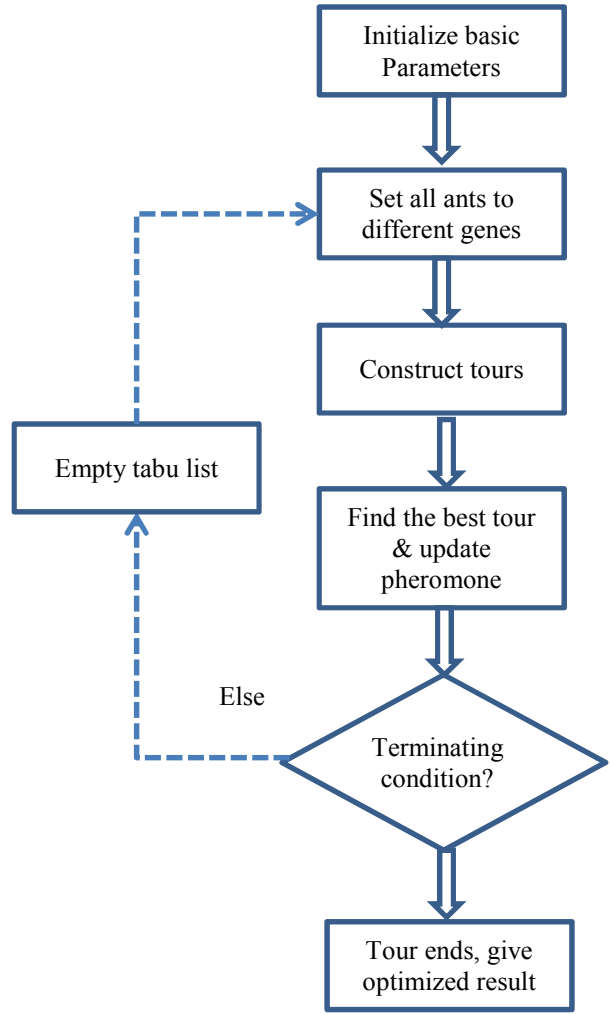

**Figure 1. Flow chart for the ACO algorithm**

The explanation of the algorithm is as follows:

i) *Conversion of gene expression to gene correlation matrix:* The expression values of the genes were converted to gene correlation matrix using Pearson correlation coefficient, as computed in [18] and a NxN matrix of correlation values were generated, where N= number of genes. Consider gene correlation matrix as a network where genes are placed to nodes and their pair-wise correlation values are assigned to edges between them.

ii) *Input:* Input for ACO algorithm is the correlation matrix of NxN genes, as computed in step 1.

iii) Maximizing the correlation function: The gene interaction network is initially represented as complete weighted graph $G=(N, E)$, where $N$ is a $n=|N|$ nodes (genes) and $E$ being the set of edges connecting the nodes (represents gene interactions). The input pairwise correlation value C is considered as weight of the edges. The correlation function (regulation function),

$$C = \sum_{i=1}^{n-1} c_{i,i+1} + c_{n,1} \qquad (2)$$

has to be maximized where $C$ is the sum of the correlation between all the gene nodes visited by ants and $c_{i,i+1}$ is the Pearson correlation between gene '$i$' and the gene '$i+1$'. Hence, we have to find da maximum correlation Hamiltonian circuit in the gene interaction network, where Hamiltonian circuit is a close walk touching each gene in the interaction network exactly one.

iv) Some basic parameters that were used for the simulation are:
   a) Ant System : Max-Min
   b) No of ants : N (No of genes)
   c) Independent Tries : 1
   d) No of tours in each trial : 100
   e) Alpha : 1.0
   f) Beta : 2.0
   g) Rho : 0.5

v) Output: The output of the simulation is the network with the maximum correlation among the $N$ genes with the maximum number of $N$ edges.

## 4. Results and Discussions

The Ant colony optimization algorithm, as described in the previous section, has been implemented in C programming language for the inference key interactions in GRN. After giving the gene correlation matrix as input to the ACO algorithm, the results were saved in a text file that stores highly correlated interactions in GRN. The first simulation was done on the SOS DNA repair genetic network dataset of e.coli generated by Ronen, et.al. [19]. This network has nearly 30 genes regulated at the transcription level. Four different experiments were conducted with different UV light intensities. Using these experiments, expressions of eight major genes, such as uvrD, uvrA, lexA, recA, umuDC, ruvA, polB and uvrY, have been documented as shown in Fig. 2. [9].

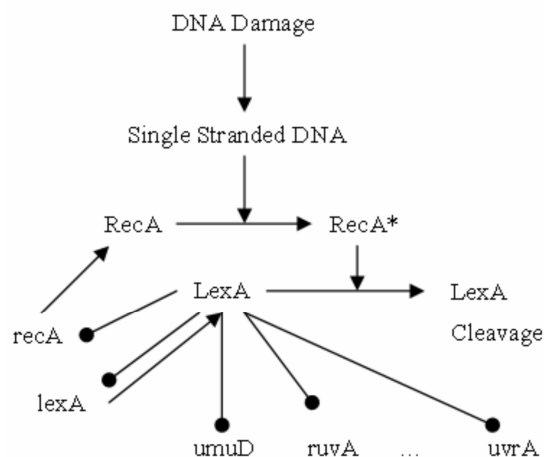

**Figure 2. SOS DNA repair network of e.coli**

The pairwise correlation between eight genes in SOS repair network has been calculated using Pearson's correlation coefficient and correlation matrix is shown in Table 1. The correlation matrix was fed as an input to the ACO algorithm. The predicted interactions and their match with the gold standard are shown in Table 2. Result shows that out of nine interactions, as reported by Ronen, et.al. [19], three interactions have been predicted correctly.

Table 1. Correlation matrix for eight genes in SOS repair network

|       | uvrD   | lexA   | umuDC  | recA   | uvrA   | uvrY    | ruvA   | polB |
|-------|--------|--------|--------|--------|--------|---------|--------|------|
| **uvrD**  | 1      |        |        |        |        |         |        |      |
| **lexA**  | 0.7647 | 1      |        |        |        |         |        |      |
| **umuDC** | 0.1982 | 0.5101 | 1      |        |        |         |        |      |
| **recA**  | 0.8013 | 0.9538 | 0.5962 | 1      |        |         |        |      |
| **uvrA**  | 0.8018 | 0.9603 | 0.4584 | 0.9779 | 1      |         |        |      |
| **uvrY**  | 0.3838 | 0.2135 | 0.2996 | 0.4535 | 0.4009 | 1       |        |      |
| **ruvA**  | 0.1912 | 0.6497 | 0.4551 | 0.5668 | 0.5796 | -0.0175 | 1      |      |
| **polB**  | 0.4326 | 0.6267 | 0.4270 | 0.6465 | 0.5855 | 0.3807  | 0.5159 | 1    |

Table 2. Predicted interactions and their match with the gold standard for all eight genes

| Gene 1 | Gene 2 | Match with gold standard |
|--------|--------|--------------------------|
| **UvrD**  | uvrA  | NO  |
| **UvrA**  | lexA  | YES |
| **LexA**  | recA  | YES |
| **RecA**  | umuDC | YES |
| **UmuDC** | ruvA  | NO  |
| **RuvA**  | polB  | NO  |
| **PolB**  | uvrY  | NO  |
| **UvrY**  | uvrD  | NO  |

In our next experiment, we used another benchmark dataset of IRMA network consisting of five genes published by Cantone, et.al., 2007 [20]. The expression values in IRMA dataset has been taken at 16 time points (at 0 minute,10 minute, 20 minute, 40 minute,…,280 minute). Out of five interactions, three interactions have been correctly identified by ACO algorithm for both kinds of data switch-on as well as switch-off, as shown in Table 3 and Table 4, respectively. The switch-on and switch-off network have two correct interactions in common and one unique correct interaction. Thus, each of them has three correct interactions. Changing various parameters in the algorithm as well as number of iterations in algorithm did not have any effect on the results. In correlation to the nature of the algorithm we inferred that the algorithm may work better with a network where relation among gene is one to one that is each gene interacts with only one gene or we can say an interacting pathway.

Table 3. Predicted interaction by for IRMA switch-on data and there match with the gold standard.

| Gene1 | Gene2 | Match with gold standard |
|-------|-------|--------------------------|
| **GAL80** | GAL4 | YES |
| **GAL4** | CBF1 | YES |
| **CBF1** | SWI5 | YES |
| **SWI5** | ASH1 | NO |
| **ASH1** | GAL4 | NO |

Table 4. Predicted interactions for IRMA switch-off data and there match with the gold standard.

| Gene1 | Gene2 | Match with gold standard |
|-------|-------|--------------------------|
| **GAL4** | GAL80 | YES |
| **GAL80** | ASH1 | NO |
| **ASH1** | CBF1 | YES |
| **CBF1** | SWI5 | YES |
| **SWI5** | GAL4 | NO |

## 5. Conclusions and Future Directions

The gene interaction network plays an important role in identifying root-cause of various diseases. The gene interaction network, especially key interactions, plays an important role in identifying biomarkers for disease that further helps in drug design. Inferring gene interaction network from gene expression data is one of the key objectives of systems biology research. Ant colony optimization is an optimization algorithm based on natural evolution and has been used in many optimization problems including some of the bioinformatics problems. This paper reports an application of ant colony optimization algorithm for inferring the key gene interactions from gene expression data. The algorithm has been tested on two different kinds of benchmark datasets, SOS DNA repair network and IRMA network. In SOS DNA repair network, out of nine interactions, three interactions have been correctly identified. In IRMA network, out of five interactions, three are correctly predicted for both kinds of data, switch-on and switch-off data. Thus, this result shows a comparatively better result over SOS DNA repair network. The limitation of proposed algorithm is that it can find out a total number of interactions equal to total number of genes. Although, the ACO-based algorithm does not produce much accuracy but results produced are robust because same result was produced after each iterations. The other advantages are that it is computationally less expensive and scalable. In the future, we may try to merge ACO-based algorithm with some other algorithm.

**Acknowledgements**

Authors would like to thank Danish Khan and Shama Aafreen for their assistance in writing ACO-based code and fruitful discussions.